\def\year{2020}\relax
\documentclass[letterpaper]{article} 
\usepackage{aaai20}  
\usepackage{times}  
\usepackage{helvet} 
\usepackage{courier}  
\usepackage[hyphens]{url}  
\usepackage{graphicx} 
\usepackage[utf8]{inputenc} 
\usepackage[T1]{fontenc}    
\usepackage{url}            
\usepackage{booktabs}       
\usepackage{amsfonts}       
\usepackage{nicefrac}       
\usepackage{microtype}      
\usepackage{amsmath}
\usepackage{algorithm}
\usepackage{algorithmic}
\usepackage{xcolor}
\usepackage{soul}
\usepackage{amssymb}
\usepackage{cases}
\usepackage{comment}
\usepackage{latexsym}
\usepackage{multirow}
\usepackage{subfig}
\usepackage{graphicx}  
\newtheorem{definition}{Definition}

\newtheorem{property}{Property}
\DeclareMathOperator*{\softmax}{softmax}
\usepackage{marvosym}
\usepackage[symbol]{footmisc}

\usepackage{hyperref}
\title{From Few to More: Large-scale Dynamic Multiagent Curriculum Learning}

\author{
Weixun Wang,\textsuperscript{\rm 1}\thanks{Equal contribution, {\dag} corresponding author.}
Tianpei Yang,\textsuperscript{\rm 1}\footnotemark[1]
Yong Liu,\textsuperscript{\rm 2}\footnotemark[1]
Jianye Hao,\textsuperscript{\rm 1,3}\footnotemark[2]\\ \Large \textbf{
Xiaotian Hao,\textsuperscript{\rm 1}
Yujing Hu,\textsuperscript{\rm 4}
Yingfeng Chen,\textsuperscript{\rm 4},
Changjie Fan,\textsuperscript{\rm 4}
Yang Gao\textsuperscript{\rm 2}}
 \\
\textsuperscript{\rm 1}{Tianjin University, \{wxwang, tpyang, jianye.hao, xiaotianhao\}@tju.edu.cn}\\
\textsuperscript{\rm 2}{Nanjing University, lucasliunju@gmail.com, gaoy@nju.edu.cn}\\
\textsuperscript{\rm 3}{Noah's Ark Lab, Huawei}\\
\textsuperscript{\rm 4}{NetEase Fuxi AI Lab, \{huyujing, chenyingfeng1, fanchangjie\}@corp.netease.com}\\
}

\begin{document}

\maketitle


\begin{abstract}
A lot of efforts have been devoted to investigating how agents can learn effectively and achieve coordination in multiagent systems. However, it is still challenging in large-scale multiagent settings due to the complex dynamics between the environment and agents and the explosion of state-action space. In this paper, we design a novel Dynamic Multiagent Curriculum Learning (DyMA-CL) to solve large-scale problems by starting from learning on a multiagent scenario with a small size and progressively increasing the number of agents. We propose three transfer mechanisms across curricula to accelerate the learning process. Moreover, due to the fact that the state dimension varies across curricula,, and existing network structures cannot be applied in such a transfer setting since their network input sizes are fixed. Therefore, we design a novel network structure called Dynamic Agent-number Network (DyAN) to handle the dynamic size of the network input. Experimental results show that DyMA-CL using DyAN greatly improves the performance of large-scale multiagent learning compared with state-of-the-art deep reinforcement learning approaches. We also investigate the influence of three transfer mechanisms across curricula through extensive simulations. \footnote[1]{We provide a video introducing our DyMA-CL and experimental demonstrations in \href{https://github.com/wwxFromTju/wwxFromTju.github.io/blob/master/video/DyMA.mp4?raw=true}{link}}
\end{abstract}


\section{Introduction}\label{sec1}
Reinforcement learning (RL) \cite{sutton} has achieved great success in achieving human-level control in complex tasks \cite{dqn}. However, there also exist a lot of challenges in multiagent systems (MASs) where a group of autonomous agents in a shared environment from which they learn what to do according to the reward signals received while interacting with each other. \cite{claus1998dynamics,bu2008comprehensive}. Furthermore, in large-scale multiagent systems, the dynamics and stochasticity of the environment become more complex, which makes it more challenging to achieve coordination among agents \cite{ic3,meanfield,jiang2018learning}. %

One efficient way to address large-scale multiagent learning problems is to leverage the concept of Curriculum Learning (CL), which has been an active field of research in the past few years, especially regarding its application to RL. The Curriculum Learning, consists of defining a set of source tasks and training the agent on each of them individually before progressing to learning on the full task. 

\begin{figure}
\centering
\includegraphics[width=\linewidth]{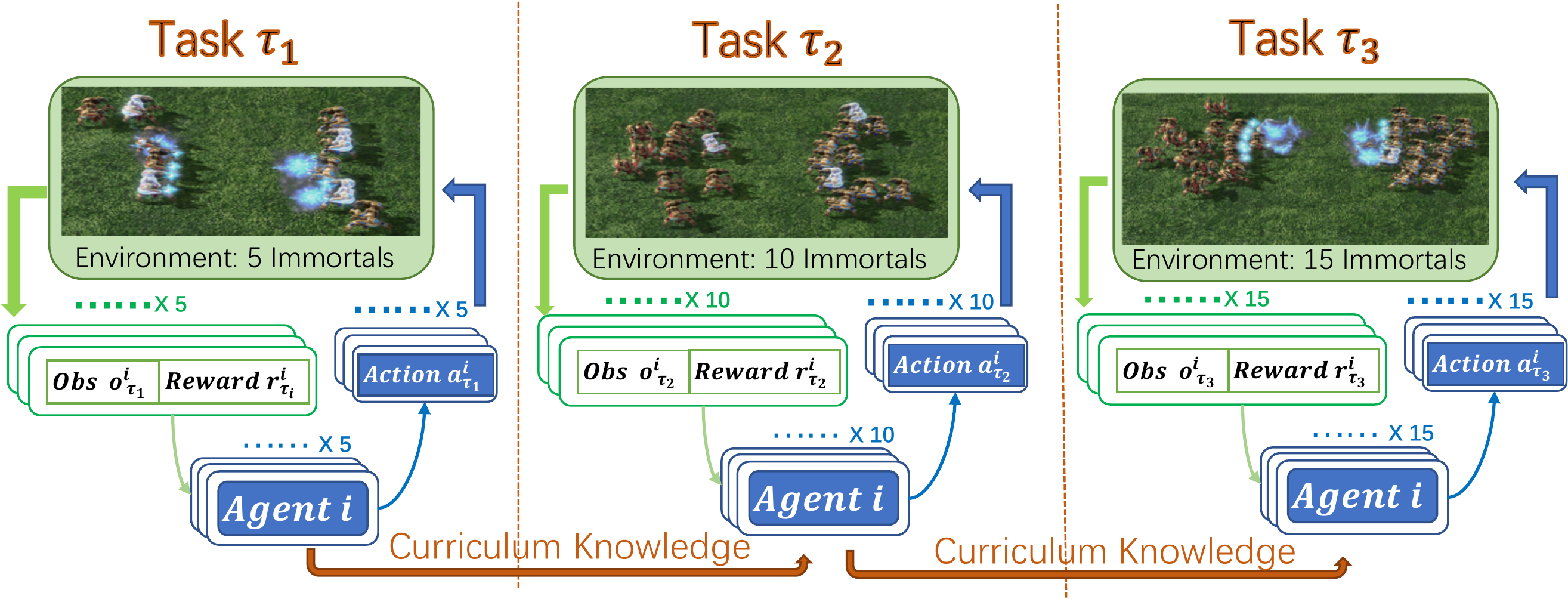}
\caption{An example of DyMA-CL in StarCraft II.} \label{fig2}
\end{figure}

One major direction of applying CL to RL focuses on how to deal with increasingly complicated tasks. Andreas et al. \shortcite{AndreasKL17} used curriculum learning to make their model scale up smoothly from simple tasks to difficult ones, the difficulty of each task is associated with sketches of different length. Later, Wu and Tian \shortcite{wutian} integrated RL with CL for the complex video game Doom, and developed an adaptive curriculum training that samples from a varying distribution of tasks to train the model, which achieves higher scores than learning the target task directly. However, these methods simply manually design the curricula which requires piror knowledge. Another direction of CL is to automatically design the curriculum. 
Narvekar et al. \shortcite{NarvekarSS17} proposed formulating the selection of tasks using a Curriculum Markov Decision Process (CMDP). However, whether the curriculum policy could actually be learned is not demonstrated. Later, they \shortcite{NarvekarS19} addressed this problem by exploring various representations to learn the curriculum policy. 

However, all the above approaches focus on designing the curricula manually or automatically in single-agent learning tasks. Although some existing works consider CL in multiagent settings, the way they utilize CL is quite simple, which is not the focus of these works \cite{mateams}. To address the growing challenges as the increase of agent-number in large-scale MASs, in this paper, we firstly propose a novel multiagent CL, named Dynamic Multiagent Curriculum Learning (DyMA-CL) as shown in Figure \ref{fig2}. DyMA-CL solves large-scale problems by starting from learning on a small-size multiagent scenario and progressively increasing the number of agents to learn the target task finally. Three kinds of transfer mechanisms (Buffer Reuse, Curriculum Distillation, and Model Reload) are proposed across different tasks to accelerate the curriculum learning process. The first two mechanisms do not require a specific network structure, while the last one does since existing network architectures cannot be directly used in such a multiagent CL setting due to the fixed size of network input and the state dimension in our settings varies across curricula. Thus, we design a novel network structure called Dynamic Agent-number Network (DyAN) by combining graph neural network to handle the dynamic size of the network input. Experimental results in Starcraft-II \cite{smac} and MAgent \cite{magent} show that DyMA-CL greatly improves the performance on large-scale problems compared with state-of-the-art DRL approaches; and three kinds of transfer mechanisms across curricula greatly boost the performance of DyMA-CL.  

\section{Background}\label{sec3}


\subsection{Partially Observable Stochastic Games}\label{sec3.1}

A natural multiagent extension of Markov decision processes (MDPs) are Stochastic Games (SGs) \cite{littman1994markov}, which model the dynamic interactions among multiple agents. In this paper, we follow previous work's settings and model the multiagent learning problems as Partially Observable Stochastic Games (POSGs) \cite{posg} considering that agents may not have access to the complete environmental information. 

A \textsl{Partially Observable Stochastic Game} (POSG) is defined as a tuple $\langle \mathcal{N}, \mathcal{S}, \mathcal{A}^1, \cdots, \mathcal{A}^n, T, \mathcal{R}^1, \cdots$,$ \mathcal{R}^n, \mathcal{O}^1, \cdots$,$ \mathcal{O}^n \rangle$, where $\mathcal{N}$ is the set of $n$ agents; $\mathcal{S}$ is the state set; $\mathcal{A}^i$ is the set of actions available to agent $i$ ($\mathcal{A} = \mathcal{A}^1\times \mathcal{A}^2 \times \cdots \times \mathcal{A}^n$ is the joint action space); $T$ is the transition function that defines transition probabilities between states: $\mathcal{S} \times \mathcal{A} \times \mathcal{S} \to \left[0,1\right]$; $\mathcal{R}^i$ is the reward function for agent $i$: $\mathcal{S} \times \mathcal{A}\to \mathbb{R}$ and $\mathcal{O}^i$ is the observation set of agent $i$. 

Note that each state $s \in \mathcal{S}$ contains the possible configurations of the environment and all agents, while each agent $i$ draws a private observation $o^i$ correlated with the state: $\mathcal{S}\mapsto \mathcal{O}^i$, e.g., an agent's observation includes the agent's private information and the relative distance between itself and other agents. Formally, an observation of agent $i$ at step $t$ can be constructed as follows: $o_t^i=\{ o_t^{i,env}, m_t^i, o_t^{i,1},\cdots,o_t^{i,i-1},o_t^{i,i+1},\cdots,o_t^{i,n} \}$, where $o_t^{i,env}$ describes the surrounding environmental information, $m_t^i$ is agent $i$'s private property (e.g., in robotics, $m_t^i$ includes agent $i$'s location, the battery power and the healthy status of each component) and the rest are the observations of agent $i$ on other agents (e.g., in robotics, $o_t^{i,i-1}$ includes the agent $i$'s observation about the relative location, the exterior of agent $i-1$).  A policy $\pi_i$: $\mathcal{O}^i\times \mathcal{A}^i \to \left[0;1\right]$ specifies the probability distribution over the action space of agent $i$. The goal of agent $i$ is to learn the optimal policy $\pi_i^*$ that maximizes the expected return with a discount factor $\gamma$: $J= \mathbb{E}_{\pi_i^*} \left[ \sum^{\infty}_{t=0}\gamma^{t}r_{t}^{i}\right]$.

%

\subsection{Curriculum Learning}\label{sec3.2}


Curriculum Learning (CL) is firstly introduced in \cite{cl} which is defined as a Machine Learning notion to improve the performance of Supervised Learning. The idea of CL is inspired by observing the way humans learn that starts with simple, small problems and gradually progresses to more complex, difficult tasks. In Curriculum Learning, the goal is to generate a series of training tasks, beginning from the simplest one and then gradually increasing the difficulty of training to improve the final asymptotic performance or decrease the training time.

Narvekar et al. \shortcite{NarvekarSLS16} firstly applied CL to RL and proposed a new CL framework. They generated a sequence of RL source tasks, named "Curriculum", trained the agent on each of the source tasks and then on the target task. Different from CL in Supervised Learning, each task in the RL curriculum is defined as an Markov Decision Process (MDP). The difficulty of each task is controlled by eliminating certain actions or states, modifying the transition or reward function, or changing the starting or terminal distributions of MDPs. The sequence of source tasks can be manually designed or automated generated  \cite{NarvekarSS17,narvekar2018learning}. In this paper, we focus on CL in multiagent RL settings and design a dynamic multiagent curriculum learning to solve large-scale multiagent learning problems.

\begin{figure*}[ht]
\centering
\includegraphics[width=\linewidth]{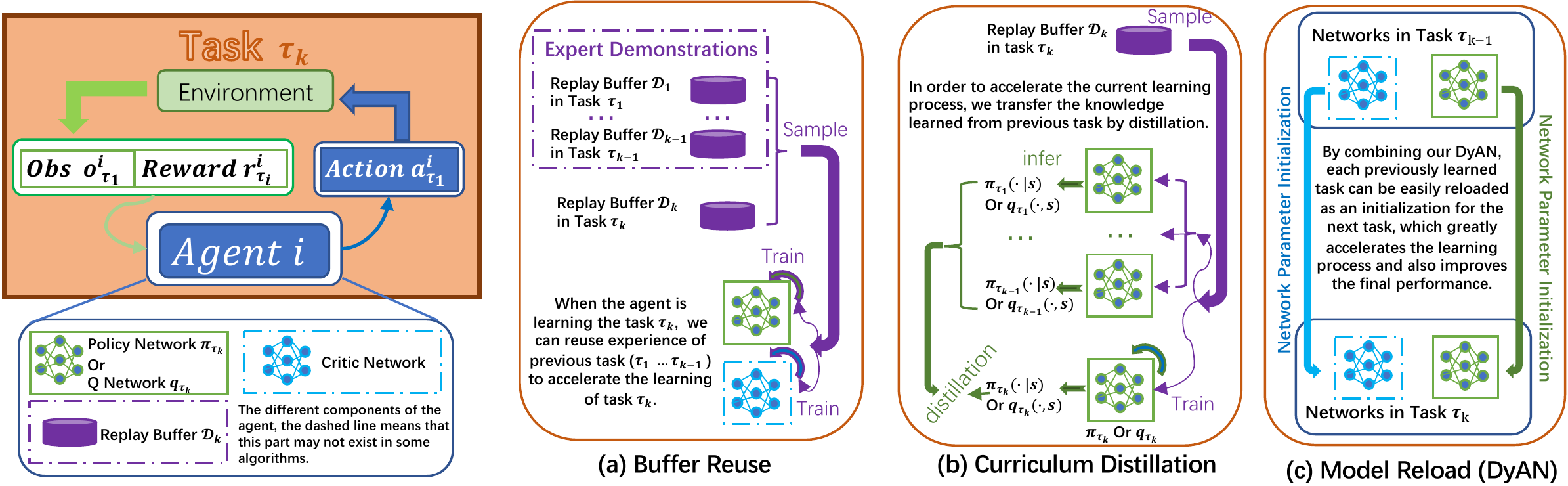}
\caption{An illustration of DyMA-CL using different transfer mechanisms.} \label{fig3}
\end{figure*}
\section{Dynamic Multiagent Curriculum Learning}\label{sec4}
\subsection{Large-scale Multiagent Systems}\label{sec4.1}
Multiagent learning receives much attention and how to achieve multiagent coordination is the key problem. Recent researches have found that the difficulty of multiagent learning is exponentially increasing as the number of agents increases \cite{smac}. Moreover, in large-scale multiagent systems, the dynamics and stochasticity of the environment become more complex, which makes it more challenging to achieve coordination among agents \cite{factorized}. We first propose several multiagent properties in nature which are commonly existing in MASs, and then utilize these properties to address large-scale multiagent learning problems.

\begin{property}\label{pro1}
Partial Observability: In MASs, agents make decisions based on their local observations, in which way large-scale problems can be reduced to relatively independent but correlated small-size ones. 
\end{property}

The common settings are to model multiagent learning problems as partially observable stochastic games (POSGs). In such partially observable environments, each agent selects an action based on its local (partial) observation, and the number of agents in each agent's vision is changing all the time as agents move and execute actions. For example, when the agent drives a car on the road, the number of cars in his local vision changes \cite{ic3}. The number of cars in the driver's vision decreased when the road is crowded at first and then some cars go out of his vision, learning in this situation is similar to learning in a scenario with a small number of cars. Therefore, large-scale learning problems can be naturally transformed into small ones based on \textit{Partial Observability}.

\begin{figure}
\centering
\includegraphics[width=\linewidth]{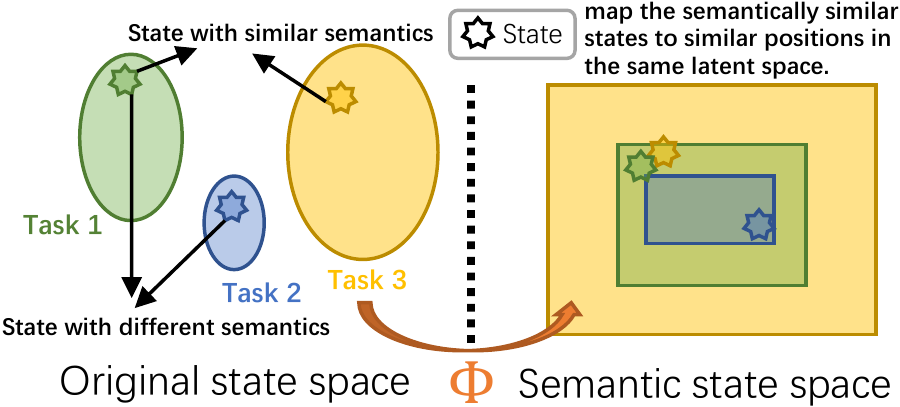}
\caption{Mapping original states to the semantic state space.} \label{figwhy}
\end{figure}

\begin{property}\label{pro2}
Sparse Interactivity: From the perspective of the global view, each agent only interacts with some of the agents in MASs at the same time, and the interactions do not happen all the time.
\end{property}

In multiagent systems, agents often only need to cooperate with their neighbors and finally achieve the overall coordination, which makes a sparse-interaction environment. For example, in the predator-prey environment \cite{YangYBWZW18}, each predator would cooperate with its neighboring predators to catch the surrounding prey, without considering other preys in a larger distance.

\begin{property}\label{pro3}
State Semanticity: Each state contains semantic information which can be utilized to measure the similarity between states.
\end{property}

Although states in different POSGs hold different dimensions, they may contain semantically similar information, which can be utilized to measure the similarity of these states. For example, in StarCraft II, with the dynamics of the game continue, the number of agents would decrease if either side of soldiers die in the battle, in which situation learning is similar to that in a small-size battlefield. As shown in Figure \ref{figwhy}, given three tasks with their state spaces in different colors respectively, we can learn a mapping function $\Phi$ to represent the relations of these states in the semantic state space. Thus, the state semanticity property can be naturally used for transfer across different multiagent scenarios.

With the increase of the number of agents, the difficulties and challenges mentioned above become more severe, which makes it harder or even impossible to learn from scratch in such large-scale multiagent systems \cite{smac}. Inspired by the above properties, we design a dynamic multiagent curriculum learning to address large-scale multiagent learning problems, i.e., starting from learning in an environment with a small number of agents, and then progressively increasing the number of agents, and finally finishing the curriculum which is described in detail in the following section.

\subsection{Knowledge Transfer across DyMA-CL}\label{sec4.2}

In this section, we propose a novel curriculum learning mechanism called dynamic multiagent curriculum learning (DyMA-CL) for efficient large-scale multiagent learning. To the best of our knowledge, it is challenging and difficult to learn on a large-scale multiagent scenario, e.g., win the battle in large-scale StartCraft II scenarios using existing methods \cite{smac,qmix}. Therefore, we build the curriculum with the increase of the agent-number to learn on a large-scale multiagent scenario. The sequence of tasks can be manually designed or automated generated. Figure \ref{fig2} illustrates an example of the DyMA-CL with 3 tasks in StarCraft II. The target task is to win on a 15 immortals vs 15 immortals (15I) scenario. We first learn the task \uppercase\expandafter{\romannumeral1} on a 5 immortals vs 5 immortals (5I) scenario, then learn the task \uppercase\expandafter{\romannumeral2} on a 10 immortals vs 10 immortals (10I) scenario and finally learn the target task. We also incorporate different knowledge transfer mechanisms across neighboring curricula which are described in detail as follows.

%
%
%

%
%

%
%




Figure \ref{fig3} shows the whole framework of DyMA-CL using different transfer mechanisms. The simplest transfer is to directly reload the model trained in previous curricula as an initialization for current task learning (Figure \ref{fig3}(c)). However, \textbf{Model Reload} is infeasible since the input of regular training networks is fixed while different curricula have different state spaces which makes the input size changing. The policy network needs to be specially designed to be suitable for different input sizes. We first propose two kinds of transfer mechanisms without any constraints on the network design: Buffer Reuse (Figure \ref{fig3}(a)) and Distillation via KL Divergence (Figure \ref{fig3}(b)). How to redesign the network to support parameter transfer will be discussed in the next section.

\textbf{Buffer Reuse} Inspired by deep Q-learning from demonstrations \cite{HesterVPLSPHQSO18} which incorporates extra expert demonstrations as the supervision, we propose a novel transfer mechanism called Buffer Reuse. For the agent learns the sequence of tasks $\tau_{1}, \tau_{2},\cdots,\tau_{k}$ using one of the off-policy RL algorithm, e.g., DQN \cite{dqn}, it is equipped with an experience replay buffer $\mathcal{D}_i$ for each task $\tau_{i}$, which stores the corresponding transition samples. When the agent is learning the task $\tau_k$, we can reuse experience of previously learned tasks to accelerate learning $c_k$. Specifically, we keep a sequence of replay buffers $\mathcal{D}_1, \mathcal{D}_2, \cdots, \mathcal{D}_{k-1}$ for previously learned tasks $\tau_{1}, \tau_{2},\cdots,\tau_{k-1}$ and sample a batch of $b$ transitions from each replay buffer equally to reuse these good transition samples as expert demonstrations, we also sample a batch from the current buffer $\mathcal{D}_k$ and then minimize the following loss in each training step:

\begin{equation}\label{eq1}
\text{Loss}=\sum_{i=1}^{k} \sum_{j=1}^{b} \left[\left(r_i^j + \gamma \max_{a_i^{'j}}q_{\tau_i}(s_i^{'j},a_i^{'j}) - q_{\tau_i}(s_i^j,a_i^{j})\right)^2\right]
\end{equation}
where $\left( s_i^j,a_i^{j},s_i^{'j},r_i^j \right)$ is the $j$th transition sample from the replay buffer $\mathcal{D}_i$ for task $i$, $\gamma$ is a discount factor.

Note that the state space is different across tasks since the number of agents varies, i.e., the dimension of a state in task $\tau_i$ is larger than that in task $\tau_j$ if $i > j$. Therefore, these samples cannot be collected together to calculate the loss in Equation (\ref{eq1}) directly. Here we modify the sampled transitions to reshape them as the same dimension first, e.g., add zero-padding for those samples with a smaller size of states, and then execute the buffer reuse mechanism to accelerate the learning process.

\textbf{Curriculum Distillation} The second transfer mechanism adopts the distillation via Kullback-Leibler (KL) divergence \cite{rusu2015policy} as the supervision which is a more general pattern suitable for both on-policy and off-policy RL algorithms.

Given a learned sequence of tasks $\mathcal{T}=\{\tau_{1}, \tau_{2}, \cdots, \tau_{k-1}\}$ and the current task $\tau_{k}$, in order to accelerate the current learning process, we transfer the knowledge from previously learned tasks by distillation. Specifically, we add an extra distillation loss $L_{\text{Distil}}$  to the regular RL loss $L_{\text{RL}}$ using the KL divergence with some temperature $\omega$: $\text{Loss} = L_{\text{RL}} + L_{\text{Distil}}$, where we can distil either Q-values or policies: 
\begin{equation}\label{kl}
\centering
\begin{aligned}
&L_{\text{Distil}}=\sum_{i=1}^{k-1}\text{KL}(\pi_{\tau_i}||\pi_{\tau_k})\quad \text{or} \\
L_{\text{Distil}}=\sum_{i=1}^{k-1}&\sum_{j=1}^{|\mathcal{D}_k|}\softmax(\frac{\mathbf{q}_{\tau_i}(s_j)}{\omega})\ln\frac{\softmax(\frac{\mathbf{q}_{\tau_i}(s_j)}{\omega})}{\softmax(\mathbf{q}_{\tau_k}(s_j))}
\end{aligned}
\end{equation}
Where, $\pi_{\tau_i}$ is the policy for task $\tau_i$ and $\omega$ is the temperature that controls the proportion transferred to the curriculum $\tau_k$. Similar to the cases in Buffer Reuse mechanism, states as the network input for different curricula should be reshaped to the same size first.

\subsection{Dynamic Number Agent Network}\label{sec4.3}

As mentioned above, each kind of transfer mechanisms cannot directly be used in our DyMA-CL, since the number of agents varies across curricula and the dimension of each agent $i$'s observation $o_t^i$ at each step $t$ changes, i.e., the number of observations for other agents changes as the number of agents changes. 

Although the state space is different in two environments due to the different number of agents, according to the Property \ref{pro3}, some states in a large-scale environment often contain similar semantic information to that in a small-size one. Therefore, we provide a formal definition of the semantic mapping function $\Phi(\cdot)$ which extracts the semantic information from each agent's observation and indicates the mapping between different state spaces.

\begin{definition}{Semantic Mapping Function}\label{def1}

Given three tasks $\tau_{e}$, $\tau_{f}$ and $\tau_{g}$ with different state dimensions, if state $s_{e}^{\tau_{e}}$ and state $s_{f}^{\tau_{f}}$ contain similar semantic information while $s_{g}^{\tau_{g}}$ does not, then through the mapping function $\Phi(\cdot)$, there exists a latent space that makes the following inequation establish: 
$\text{dis}(\Phi(s_{e}^{\tau_e}),\Phi(s_{f}^{\tau_f})) < \text{dis}(\Phi(s_{e}^{\tau_e}),\Phi(s_{g}^{\tau_g}))$, $\text{dis}(\Phi(s_{e}^{\tau_e}),\Phi(s_{f}^{\tau_f})) < \text{dis}(\Phi(s_{f}^{\tau_f}),\Phi(s_{g}^{\tau_g}))$, 
where $dis(,)$ is the distance between two vectors.
\end{definition}

By the definition of Semantic Mapping Function $\Phi(\cdot)$, the states in each task $\tau_{i}$ (each POSG) can be transformed into the same semantic state space, which is also suitable for mapping the local observation of each agent to the same latent space. Thus, we can transfer knowledge across POSGs with different state dimensions. This concept is widely used in domain adaptation area \cite{HigginsPRMBPBBL17,ArnekvistKS19}, while they focus on how to transfer from different tasks with the same state dimension. However, the biggest challenge for DyMA-CL is how to deal with different network input dimensions caused by the different number of agents, and map the semantically similar states to similar positions in the same latent space.

If the network is not restricted by the state/observation of different dimensions, or the states/observations with the same semantics of different dimensions can be mapped to similar positions in the same latent space, then we can easily transfer knowledge from different numbers of POSGs using any of the above mechanisms for efficient curriculum learning. Inspired by Graph Neural Network (GNN) \cite{XuHLJ19}, to this end, we propose a novel network architecture named \textbf{Dy}namic \textbf{A}gent-number \textbf{N}etwork (DyAN) to address the above problems.

\begin{figure}
\centering
\includegraphics[width=0.9\linewidth]{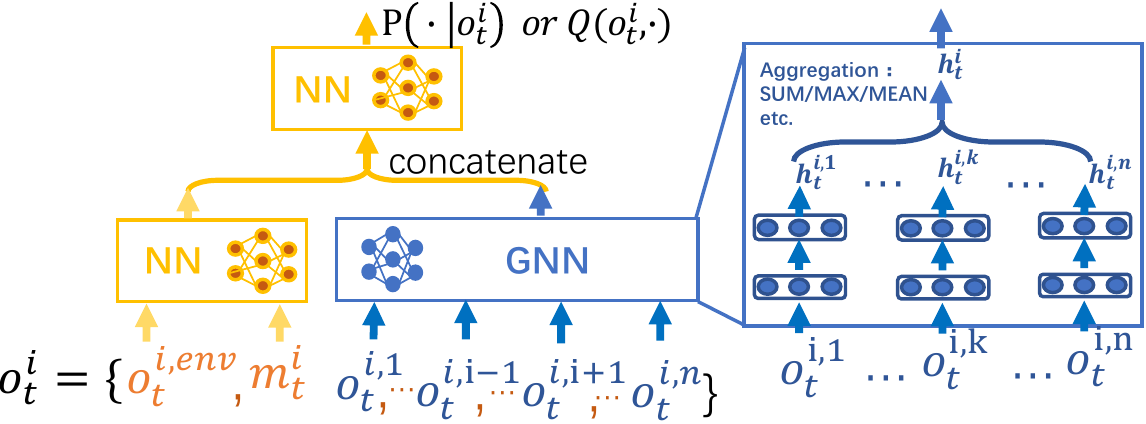}
\caption{The network structure of DyAN.} \label{gnn1}
\end{figure}

Figure \ref{gnn1} shows the network structure of the DyAN. Given an observation $o_t^i$ of agent $i$ at step $t$, the left part of DyAN is the general neural networks, e.g., the fully-connected layers, with the environmental information $o_t^{i,env}$ and its private property $m_t^i$ as input. While the reset of $o_t^i$ contains several observations about other agents which change among different curricula. The right part of DyAN incorporates the GNN to handle this dynamic dimensions of input. Specifically, we learn a representation $h_t^{i,j}$ for the agent $i$'s observation $o_t^{i,j}$ on each other agent $j$, which is achieved after several neural network layers; and then using an aggregation operator to get the output of GNN. Formally, the output of a GNN is:
\begin{equation}\label{eqgnn}
h_{t}^{i} = \text{AGGREGATE} \left( \lbrace h_{t}^{i,j}: j \in N^{-i} \rbrace \right)
\end{equation} 
where, $N^{-i}$ is the set of agents excluding agent $i$. Note that we use one layer of GNN, a multiple layers of a GNN with neighborhood communication is also suitable here. There are several alternatives for the $\text{AGGREGATE}$ operator \cite{XuHLJ19}, e.g., the $\text{MAX}$ operator that represents an element-wise max-pooling, the $\text{MEAN}$ operator representing an element-wise mean pooling and the $\text{SUM}$ operator, which performance is investigated in the following section. Next, the outputs of two parts of DyAN are concatenated to input to the following neural network layers. The final output is the Q-values or the policy respectively which is subject to the specific RL algorithms. 

As we described earlier, the simplest transfer mechanism of model reload cannot be directly used in our curriculum learning due to the dynamic dimensions of the network input. By combining our DyAN, each previously learned model can be easily reloaded as an initialization for the next curriculum learning, which greatly accelerates the learning process and also improves the final performance. In the next section, we investigate the performance of three transfer mechanisms in our curriculum learning in detail.


\section{Simulations}\label{sec5}

\begin{figure*}[ht]
\centering
\includegraphics[width=0.85\linewidth]{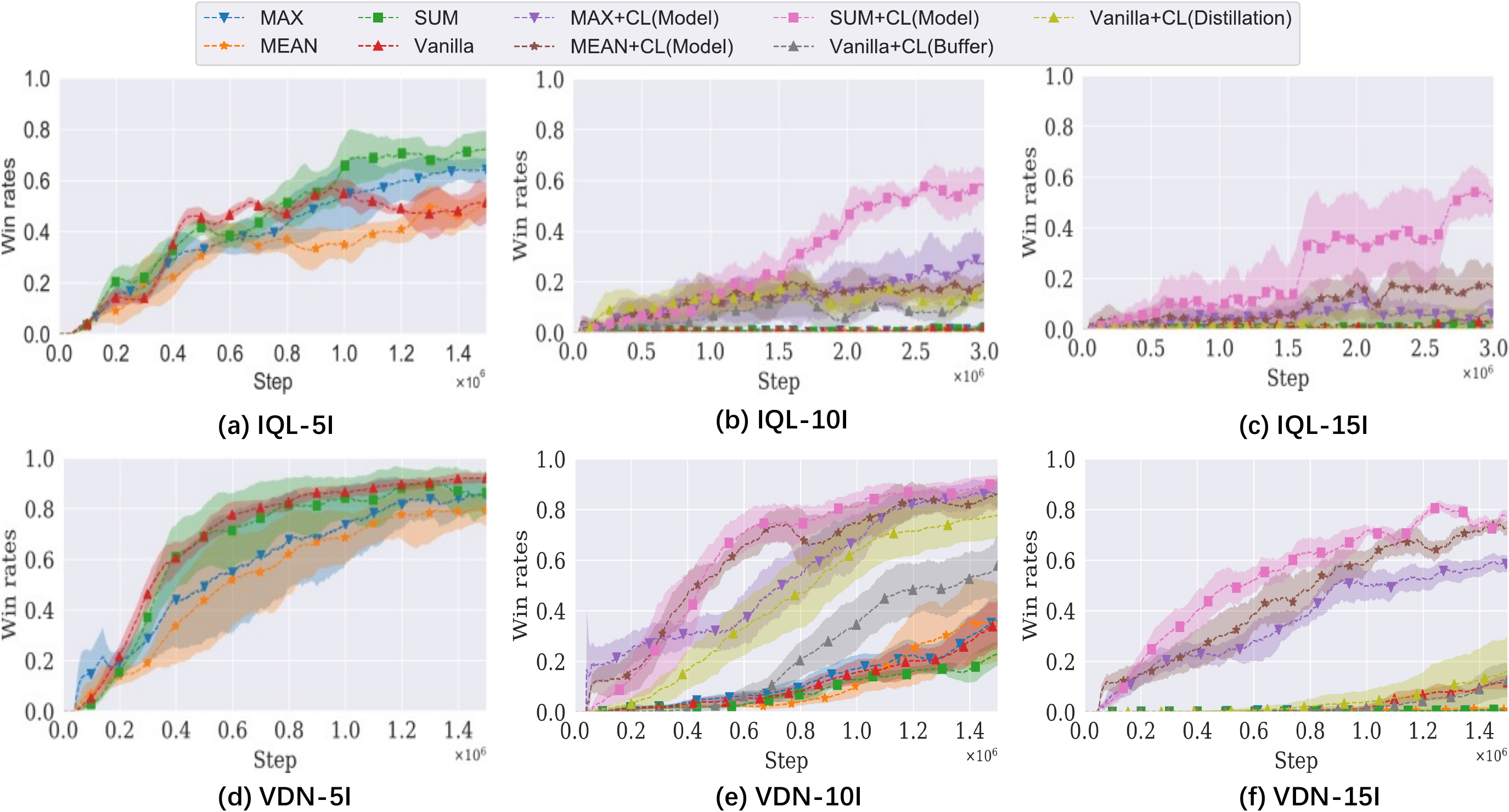}
\caption{Average win rate of IQL and VDN on DyMA-CL.}\label{iql}
\end{figure*}

In this section, we evaluate the performance of our DyMA-CL on two large-scale scenarios: 1) StarCraft II, which contains various scenarios for a number of agents to learn coordination to solve complex tasks; and 2) MAgent \cite{magent}, which is a simulated battlefield with two large-scale armies (groups), e.g., each army consists of 50 soldiers who would be arrayed in the battlefield (a grid world). We first select two representative DRL algorithms from the perspective of Independent learning and Joint-action Learning respectively: IQL \cite{iql}, VDN \cite{vdn} to investigate the performance of these approaches with and without DyMA-CL on large-scale StarCraft II scenarios. We further compare the performance of various existing DRL approaches (IQL, PPO \cite{ppo}, A2C \cite{a2c}, and ACER \cite{BHMMKF17}) with DyMA-CL on large-scale MAgent scenarios to validate the performance of DyMA-CL since independent learning is more difficult to learn in such large-scale multiagent settings without considering the coexistence of other agents. The details of neural network structures, parameter settings and the curriculum schedule are in the supplementary materials.

\subsection{StarCraftII}\label{sec5.2}
StarCraft II is a real-time strategy game with one or more humans competing against each other or a built-in game AI. At each step, each agent observes the local game state which consists of the following information for all units in its field of view: relative distance between other units, the position and unit type (detailed in supplementary materials) and selects one of the following actions: move north, south, east or west, attack one of the grid units, stop and the null action. Agents belonging to the same side receive the same joint reward at each time step that equals to the total damage on the enemy units. Agents also receive a joint reward of 10 points after killing each opponent, and 200 points after killing all opponents. The game ends when all agents on one side die or the time exceeds a fixed period. 

Note that previous StarCraft II settings enable an agent to attack one of its enemies by choosing one of id numbers \cite{smac}. In this paper, we design the attack action is to choose one of the grid units by dividing the battlefield into several grids, in which case the coordination among agents is much more difficult to achieve. We mainly consider combat scenarios and design a multiagent curriculum learning with the number of agents increasing (see Figure \ref{fig2}) to achieve the victory on a 15I scenario.

Figure\ref{iql}(a-c) show the average win rate of IQL with and without our DyMA-CL under different network structures (i.e., Vanilla network does not contain the GNN part, and MAX means the GNN uses MAX as the aggregation operator). We can see from Figure\ref{iql}(a) that SUM performs better than other kinds of network structures on the first task of a 5I scenario. As for the task \uppercase\expandafter{\romannumeral2} on a 10I battlefield (Figure\ref{iql}(b)), our DyMA-CL with all three transfer mechanisms perform better than learning from scratch. Note that the model reload mechanism performs best among all transfer mechanisms, this is because our proposed DyAN successfully learns the similar semantics of states across curricula, then the model from previously learned curriculum can be directly reused, which leads to a higher win rate. The SUM operator performs best among all three aggregations which means the capability of learning state semantics is different for these three aggregations and the GNN with SUM as aggregation learns more accurate state semantics. This will be explained in detail in the following section. For our last curriculum (Figure\ref{iql}(c)), our DyMA-CL with model reload mechanism performs best among all transfer mechanisms. Similar results as in curriculum \uppercase\expandafter{\romannumeral2} can be found that learning from scratch is too difficult as the increase of the agent number, and the winning rate is never increased. We have conducted the simulation on the same amount of total training time ($7.5e+6$ steps) and vanilla IQL still achieves an average win rate of 0.

\begin{figure}[ht]
\centering
\includegraphics[width=0.8\linewidth]{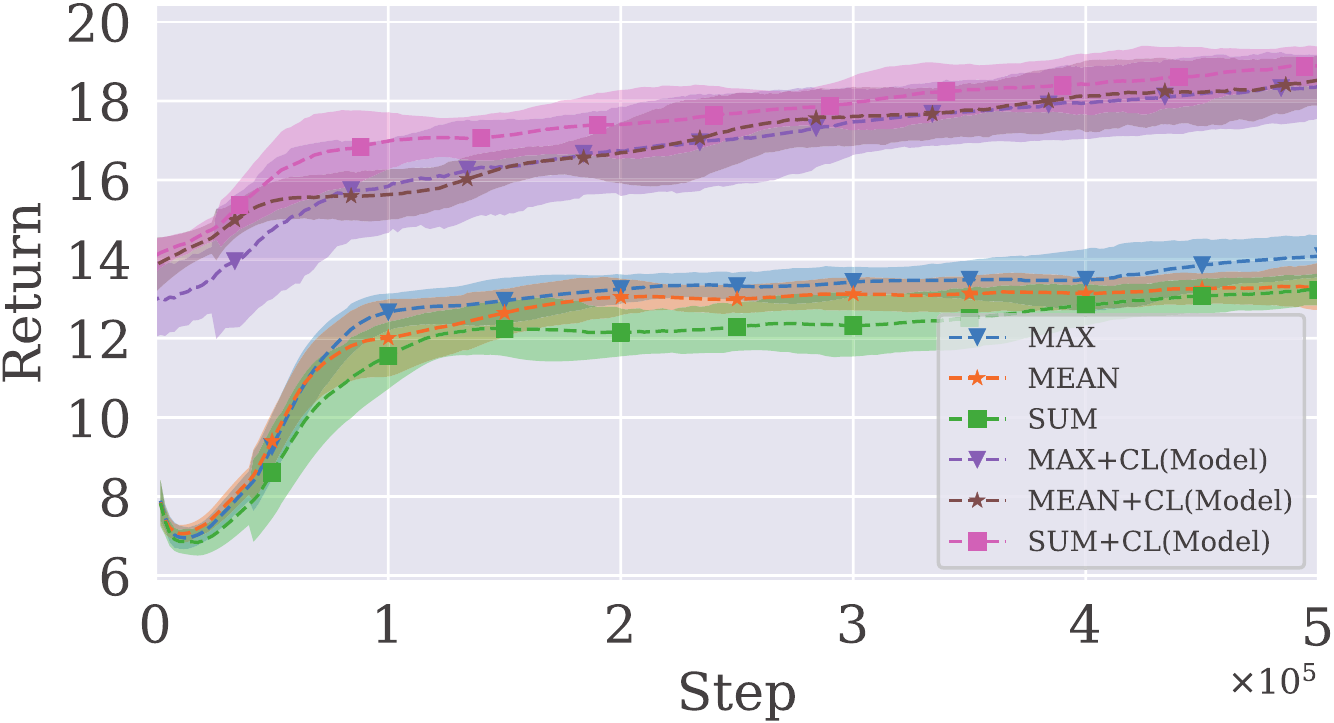}
\caption{The performance of VDN on a 15I scenario.} \label{jump}
\end{figure}

Figure\ref{iql}(d-f) depict the average win rate of VDN with and without our DyMA-CL. We can find different network architectures perform similarly on the first task learning (Figure\ref{iql}(d)), and perform better than that combining IQL in universal. This is because VDN explicitly considers how to coordinate multiple agents using a team reward. Figure\ref{iql}(e) and (f) show the similar and more outstanding performance of DyMA-CL than that in IQL, and the GNN with SUM as aggregation learns best among all three mechanisms of DyMA-CL, and the reason will be discussed in the following section in detail. Note that the common measurement for StarCraft II is the average win rate \cite{smac}, which may hinder the phenomenon of a jumpstart on the performance of DyMA-CL with model reload mechanism. Therefore, we further present the results of average rewards as shown in Figure \ref{jump}. We can see a jumpstart average reward of our DyMA-CL with model reload mechanism than learning from scratch, which indicates the agents can kill more enemies and achieve higher average rewards at the beginning than learning from scratch, confirming the effectiveness of model reload mechanism across curricula. 

\begin{figure*}
\centering
\includegraphics[width=0.85\linewidth]{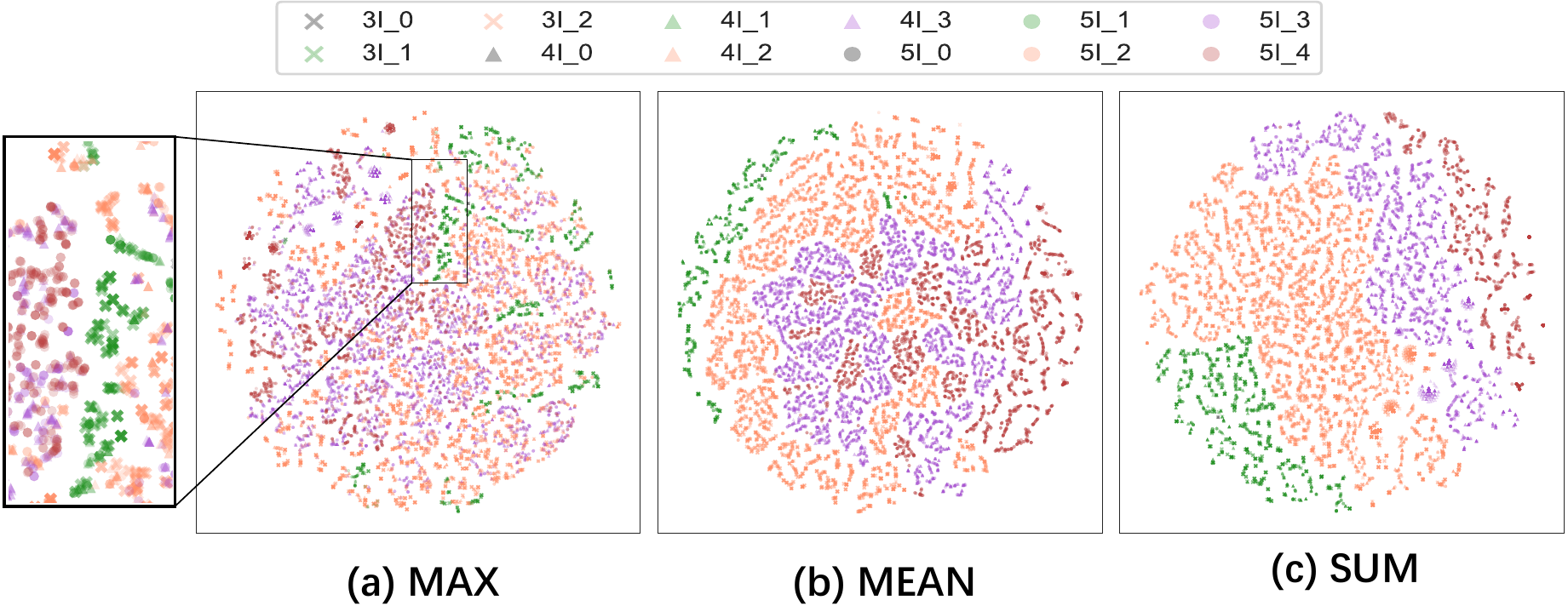}
\caption{Embedding analysis for different aggregation mechanisms.} \label{vdn_analysis}
\end{figure*}

\subsubsection{Analysis}\label{sec5.3}
We further investigate the influence of different aggregations on the performance of DyMA-CL. As we discussed in Definition \ref{def1}, if the two states with different dimensions contain similar semantic information, we can map them to the same neighborhood position in the same latent space. Here we illustrate whether these aggregations learn the semantics of states using three StartCraft II scenarios as examples, each of which contains two groups of 3, 4, 5 agents respectively. Then we input the observation about teammates to DyAN, and use t-SNE \cite{wattenberg2016use} to map the embedding output of the GNN part to a 2-dimension space, as shown in Figure \ref{vdn_analysis}(a-c). The different colors denote the state contains different semantic information, e.g., the green color represents that the local observation only contains one teammate, which is actually the same semantics while the input size is different across 3 scenarios. The different shapes represent states in different scenarios, e.g., the triangle denotes the observations from a 4I scenario. We can see that the mapping result on the SUM aggregation is best among all aggregations, which means SUM learns more accurate state semantics so that the states with the same semantics across different scenarios are mapped to the similar position and each kind of semantics is distinguished clearly. Thus this explains why SUM performs best among three aggregation operations shown in Figure \ref{iql}.

\subsection{MAgent}\label{sec5.1}
\begin{table}[ht]
\centering
\fontsize{8}{10}\selectfont
\caption{Mean and Standard Error in MAgent ('w/' denotes with and 'w/o' denotes without).} \label{table1}
\begin{tabular}{|l|c|c|c|c|c|c|}
\hline
\multicolumn{3}{|l|}{Results / Methods}  & Survivors & Kill count \\ 
\hline
\multicolumn{1}{|l|}{\multirow{6}{*}{IQL}} & \multicolumn{1}{c|}{\multirow{2}{*}{Max}}  & w/ CL    & 20.35$\pm$4.87    & 50$\pm$0  \\ 
\cline{3-5} 
\multicolumn{1}{|l|}{}  & \multicolumn{1}{c|}{}  & w/o CL & 0.54$\pm$2.65 & 31.83$\pm$8.52 \\ 
\cline{2-5} 
\multicolumn{1}{|l|}{} & \multicolumn{1}{c|}{\multirow{2}{*}{Mean}} & w/ CL    & 2.33$\pm$2.9 & 43.58$\pm$8.29      \\ 
\cline{3-5} 
\multicolumn{1}{|l|}{}& \multicolumn{1}{c|}{}& w/o CL & 0                                 & 5.37$\pm$6.92     \\ 
\cline{2-5} 
\multicolumn{1}{|l|}{}  & \multicolumn{1}{c|}{\multirow{2}{*}{Sum}}  & w/ CL    & 10.52$\pm$6.27 & 49.34$\pm$ 2.23      \\ 
\cline{3-5} 
\multicolumn{1}{|l|}{}  & \multicolumn{1}{c|}{}  & w/o CL & 0.05$\pm$ 0.42        & 26.01$\pm$ 7.11 \\ 
\hline
\multicolumn{1}{|l|}{\multirow{6}{*}{PPO}} & \multicolumn{1}{c|}{\multirow{2}{*}{Max}}  & w/ CL &  0.21$\pm$1.32 & 22.76$\pm$9.06 \\ 
\cline{3-5} 
\multicolumn{1}{|l|}{}  & \multicolumn{1}{c|}{}  & w/o CL & 0 & 4.18$\pm$4.84 \\ 
\cline{2-5} 
\multicolumn{1}{|l|}{} & \multicolumn{1}{c|}{\multirow{2}{*}{Mean}} & w/ CL  & 0.22$\pm$0.95 &  23.66$\pm$7.52\\ 
\cline{3-5} 
\multicolumn{1}{|l|}{}& \multicolumn{1}{c|}{}& w/o CL & 0 &0.34$\pm$0.61  \\ 
\cline{2-5} 
\multicolumn{1}{|l|}{}  & \multicolumn{1}{c|}{\multirow{2}{*}{Sum}}  & w/ CL &1.48$\pm$2.69 & 38.25$\pm$7.33\\ 
\cline{3-5} 
\multicolumn{1}{|l|}{}  & \multicolumn{1}{c|}{}  & w/o CL & 0 & 1.06$\pm$1.12\\ 
\hline
\multicolumn{1}{|l|}{\multirow{6}{*}{A2C}} & \multicolumn{1}{c|}{\multirow{2}{*}{Max}}  & w/ CL &16.5$\pm$10.92 &47.71$\pm$5.61  \\ 
\cline{3-5} 
\multicolumn{1}{|l|}{}  & \multicolumn{1}{c|}{}  & w/o CL &3.65$\pm$5.58 &36.4$\pm$13.79  \\ 
\cline{2-5} 
\multicolumn{1}{|l|}{} & \multicolumn{1}{c|}{\multirow{2}{*}{Mean}} & w/ CL &6.8$\pm$7.39 &43.77$\pm$10.36  \\ 
\cline{3-5} 
\multicolumn{1}{|l|}{}& \multicolumn{1}{c|}{}& w/o CL &0.28$\pm$1.04  &25.21$\pm$11.7  \\ 
\cline{2-5} 
\multicolumn{1}{|l|}{}  & \multicolumn{1}{c|}{\multirow{2}{*}{Sum}}  & w/ CL &8.96$\pm$8.52 &44.59$\pm$9.17 \\ 
\cline{3-5} 
\multicolumn{1}{|l|}{} & \multicolumn{1}{c|}{} & w/o CL &1.07 $\pm$3.75 & 17.1$\pm$14.44 \\ 
\hline
\multicolumn{1}{|l|}{\multirow{6}{*}{ACER}} & \multicolumn{1}{c|}{\multirow{2}{*}{Max}} & w/ CL  &0 &9.14$\pm$4.52  \\ 
\cline{3-5} 
\multicolumn{1}{|l|}{}  & \multicolumn{1}{c|}{}  & w/o CL &0 &4.19$\pm$2.52  \\ 
\cline{2-5} 
\multicolumn{1}{|l|}{} & \multicolumn{1}{c|}{\multirow{2}{*}{Mean}} & w/ CL  &0 &12.62$\pm$3.68  \\ 
\cline{3-5} 
\multicolumn{1}{|l|}{}& \multicolumn{1}{c|}{}& w/o CL &0 &4.84$\pm$2.8 \\ 
\cline{2-5} 
\multicolumn{1}{|l|}{}& \multicolumn{1}{c|}{\multirow{2}{*}{Sum}}& w/ CL &0 &9.67$\pm$3.68   \\ 
\cline{3-5} 
\multicolumn{1}{|l|}{}  & \multicolumn{1}{c|}{}  & w/o CL &0   &4.84$\pm$2.68  \\ 
\hline
\end{tabular}
\end{table}
MAgent is a Mixed Cooperative-Competitive scenario with two armies fighting against each other, which supports hundreds to millions of agents. The goal of each army is to get more rewards by collaborating with teammates to destroy all opponents. Each agent selects one of the following actions: moving to some grid unit or attacking some grid unit based on its local observation which contains the following information for all units: the hit points (HP), the positions. We adopt the default reward setting: -0.005 for every move, 0.2 for attacking an enemy, 5 for killing an enemy, -0.1 for attacking an empty grid, and -0.1 for being attacked or killed. We design a curriculum containing 5 tasks, each of which learns on a battlefield with different number of agents ($10 \text{vs} 10,20 \text{vs} 20,30 \text{vs} 30,40 \text{vs} 40,50 \text{vs} 50$).

We validate the performance of various independent learning algorithms with or without DyMA-CL. Table \ref{table1} presents the average survival teammates and kill count of various approaches in the target task of a 50 agents vs 50 agents scenario. We can see that DyMA-CL with model reload mechanism greatly improves the final performance of IQL than learning from scratch, achieving more survival teammates and a higher average kill count. Similar results can be found in PPO, A2C, and ACER that DyMA-CL boosts the performance of these approaches and outperforms learning from scratch. Note that the performance of ACER is worse than other methods, which is caused by the policy adjustment using samples from its replay buffer. This mechanism is only considered from the perspective of independent learning, ignoring the non-stationary environment caused by other agents. Moreover, DyMA-CL still improves the performance of ACER and achieves a higher average kill count.

\section{Discussion}
As noted, we manually design the curriculum for both two domains, StarCraft II and MAgent. Experimental results have shown the great improvement of DyMA-CL on large-scale MASs. However, the boost in the performance in this paper is the first step that validates the effectiveness of DyMA-CL. We have found that the design of the curriculum is a critical factor in the performance of DyMA-CL. How to select an appropriate curriculum schedule (including how to decide on the training step-size for each task, how to select the suitable learned model to reload and so on) is crucial. Researches about automatic generation of the curriculum are still investigated at an initial stage and not considered in multiagent settings. Perhaps the major remaining limitations are how to automatically generate the multiagent curriculum, which will be further investigated as our future work.

\section{Conclusion and Future Work}\label{sec6}
In this paper, we propose a novel algorithm, Dynamic Multiagent Curriculum Learning (DyMA-CL) to address large-scale multiagent learning problems. We also propose three transfer mechanisms across different curricula to accelerate the learning process, which is extensively validated by simulations. Furthermore, we design a novel network structure, Dynamic Agent-number Network (DyAN) to handle the dynamic size of network input. Experimental results show that DyMA-CL greatly improves the performance in large-scale problems compared with state-of-the-art DRL approaches. As future work, it is worthwhile investigating how to achieve automatically multiagent curriculum learning to accelerate large-scale multiagent learning. Another direction is how to design more efficient transfer mechanisms to facilitate robust multiagent curriculum learning.

\section{Acknowledgements}
This work is supported by Science and Technology Innovation 2030 - ``New Generation Artificial Intelligence'' Major Project No.(2018AAA0100905), and the National Natural Science Foundation of China (Grant Nos.: 61702362, 61432008, U1836214).
\bibliographystyle{aaai}
\bibliography{DyAN}

\section*{Supplementary Materials}
\subsection{Experimental Description}

\subsubsection{StarCraft II} 
In StarCraft II, we follow the settings of previous works \cite{qmix,smac}. The local observation of each agent is drawn within their field of view, which encompasses the circular area of the map surrounding units and has a radius equal to the sight range. The input vector of each agent consists of the following features for all units in its observation range (both teammates and enemy): distance, relative x, relative y, and the unit type. We design the curriculum that each agent first learns on a 5 Immortals vs 5 Immortals battlefield for $1.5e+6$ steps, then learns on a 10 Immortals vs 10 Immortals battlefield for $1.5e+6$ steps, and learns on the target task of a 15 Immortals vs 15 Immortals battlefield for $1.5e+6$ steps. We add $1.5e+6$ training steps for IQL in last two tasks since it is more difficult to learn in such large-scale multiagent settings without considering the coexistence of other agents.

\subsubsection{MAgent} 
\begin{figure}
\centering
\includegraphics[width=\linewidth]{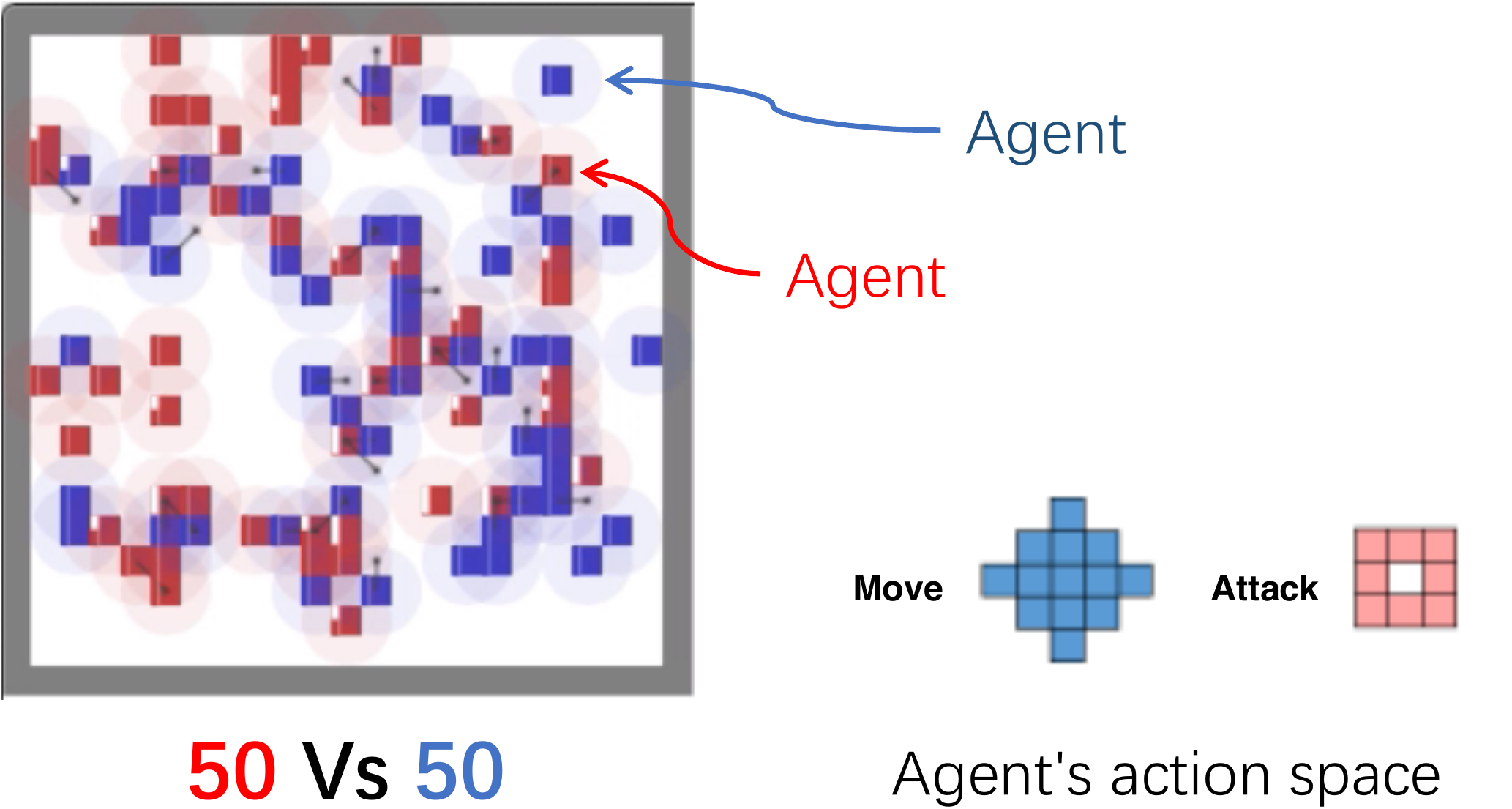}
\caption{An illustration of action range on MAgent.} \label{action}
\end{figure}

We following the settings of previous work on MAgent \cite{magent}, the action space includes 13 move actions, each of which will leads to a corresponding direction; and 8 attack actions, each of which attacks a corresponding grid unit (see Figure \ref{action}). The observation range for each agent is a 13 $\times$ 13 range of grids. The input vector of each agent includes the position and the Hit Point (HP) of the agent; the relative position of teammates and enemies, which is represented as a one-hot vector; the Hit Points of teammates and enemies, the number of teammates and enemies; the action, reward and normalized position of the agent at previous step. We design the curriculum that each agent learns the sequence of tasks as follows: learning on a 10 agents vs 10 agents battlefield for $7500$ steps; learning on a 20 agents vs 20 agents battlefield for $4500$ steps; learning on a 30 agents vs 30 agents battlefield for $1500$ steps; learning on a 40 agents vs 40 agents battlefield for $750$ steps; learning the target task of a 50 agents vs 50 agents battlefield for $1e+4$ steps. 

\begin{figure*}
\centering
\includegraphics[width=\linewidth]{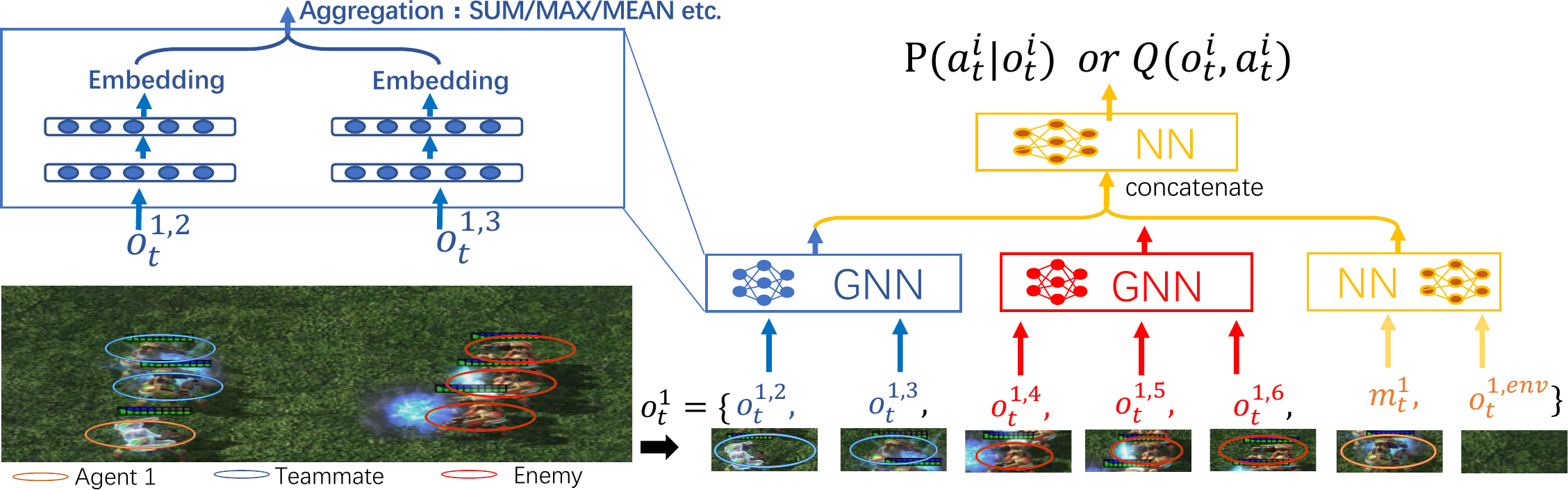}
\caption{An illustration of the network structure of DyAN for StarCraft II.} \label{dyan}
\end{figure*}

An illustration of DyAN for StarCraft II is shown in Figure \ref{dyan}. For the vanilla network which does not contain the right part of DyAN, it contains a fully-connected layer with 64 units, following a GRU layer with 64 units, and then an output layer that outputs the state-action values of each action.
Our DyAN divides the observation of each agent into two parts, the environmental information $o_t^{1,env}$ and itself information $m_t^{1}$ are input to a fully-connected layer with 64 units; the rest of information is input to the right part of DyAN. Specifically, we separate the observation for its teammates ($o_t^{1,2},o_t^{1,3}$) and enemies ($o_t^{1,4},o_t^{1,5},o_t^{1,6}$), and input to two fully-connected layers with 64 units respectively, each of which follows an aggregation function, then all three parts are concatenated together and input to a GRU layer with 64 units, the output layer is a fully-connected layer that outputs the state-action values of each action. 

\subsection{Network Structure}
\subsubsection{StarCraft II} 

\subsubsection{MAgent}  The network structure of DyAN for MAgent is similar to that for StarCraft II, except that the unit size for each neural network layer is 16. The output can be either state-action values for each action, or the probability of choosing each action through a $\text{SOFTMAX}$ activation. For actor-critic approaches, e.g., A2C \cite{a2c}, it also outputs the state-values.

\subsection{Parameter Settings}
Here we provide the hyperparameters for StarCraft II and MAgent.

\begin{table}[ht]
\caption{Hyperparameters used for StarCraft II.}
\centering
\begin{tabular}{c|c}
\rule{0pt}{12pt}
 Hyperparameter & Value\\
 \hline
 \rule{0pt}{12pt}
 Batch-size & 32 \\
 \rule{0pt}{12pt}
 Replay memory size &5000 \\
 \rule{0pt}{12pt}
 Discount factor($\gamma$) & 0.99\\
 \hline
 \rule{0pt}{12pt}
 Optimizer & RMSProp \\
 \rule{0pt}{12pt}
 Learning rate& $5e-4$ \\
 \rule{0pt}{12pt}
 $\alpha$ & 0.99 \\
 \rule{0pt}{12pt}
 $e$& $1e-5$\\
 \rule{0pt}{12pt}
 Gradient-norm-clip& 10 \\
\hline
 \rule{0pt}{12pt}
Action-selector&$\epsilon$-greedy\\
\rule{0pt}{12pt}
$\epsilon$-start& 1.0\\
\rule{0pt}{12pt}
$\epsilon$-finish& 0.05\\
\rule{0pt}{12pt}
$\epsilon$-anneal-time& 50000 step\\
\hline
 \rule{0pt}{12pt}
target-update-interval& 200\\
\hline
\end{tabular}
\end{table}

\begin{table}[ht]
\caption{IQL hyperparameters used for MAgent.}
\centering
\begin{tabular}{c|c}
\rule{0pt}{12pt}
 Hyperparameter & Value\\
 \hline
 \rule{0pt}{12pt}
 Batch-size & 32 \\
 \rule{0pt}{12pt}
 Replay memory size &100000 \\
  \rule{0pt}{12pt}
 Replay memory size at the start of training &5000 \\
 \rule{0pt}{12pt}
 Discount factor($\gamma$) & 0.98\\
 \hline
 \rule{0pt}{12pt}
 Optimizer & Adam \\
 \rule{0pt}{12pt}
 Learning rate& $1e-4$ \\
 \rule{0pt}{12pt}
 $e$ & $1e-8$ \\
\hline
 \rule{0pt}{12pt}
Action-selector&$\epsilon$-greedy\\
\rule{0pt}{12pt}
$\epsilon$-start& 1.0\\
\rule{0pt}{12pt}
$\epsilon$-finish& 0.01\\
\rule{0pt}{12pt}
$\epsilon$-anneal-time& 99 episodes\\
\hline
 \rule{0pt}{12pt}
target-update-interval& 20\\
\hline
\end{tabular}
\end{table}

\begin{table}[ht]
\caption{A2C hyperparameters used for MAgent.}
\centering
\begin{tabular}{c|c}
\rule{0pt}{12pt}
 Hyperparameter & Value\\
 \hline
 \rule{0pt}{12pt}
Training interval (T horizon) & 20 step\\
 \rule{0pt}{12pt}
 Discount factor($\gamma$) & 0.98\\
 \hline
 \rule{0pt}{12pt}
 Optimizer & Adam \\
 \rule{0pt}{12pt}
 Learning rate& $1e-3$ \\
 \rule{0pt}{12pt}
 $e$ & $1e-8$ \\
\hline
\rule{0pt}{12pt}
Entropy term coefficient & 0.1 \\
\rule{0pt}{12pt}
Value loss coefficient& 1 \\
\rule{0pt}{12pt}
Actor loss coefficient& 1 \\
\hline
\end{tabular}
\end{table}

\begin{table}[ht]
\caption{PPO hyperparameters used for MAgent.}
\centering
\begin{tabular}{c|c}
\rule{0pt}{12pt}
 Hyperparameter & Value\\
 \hline
 \rule{0pt}{12pt}
Training interval (T horizon) & 10 step\\
 \rule{0pt}{12pt}
 Discount factor($\gamma$) & 0.98\\
 \hline
 \rule{0pt}{12pt}
 Clip hyperparameter $\epsilon$ & 0.2 \\
  \rule{0pt}{12pt}
  GAE $\lambda$ & 0.95 \\
 \hline
 \rule{0pt}{12pt}
 Optimizer & Adam \\
 \rule{0pt}{12pt}
 Learning rate& $2e-3$ \\
 \rule{0pt}{12pt}
 $e$ & $1e-8$ \\
\hline
\rule{0pt}{12pt}
Entropy term coefficient & 1e-3 \\
\rule{0pt}{12pt}
Value loss coefficient& 1 \\
\rule{0pt}{12pt}
Actor loss coefficient& 1 \\
\hline
\end{tabular}
\end{table}

\begin{table}[ht]
\caption{ACER hyperparameters used for MAgent.}
\centering
\begin{tabular}{c|c}
\rule{0pt}{12pt}
 Hyperparameter & Value\\
 \hline
 \rule{0pt}{12pt}
 Batch-size & 64 \\
 \rule{0pt}{12pt}
 Replay memory size &100000 \\
  \rule{0pt}{12pt}
 Replay memory size at the start of training &100 \\
 \rule{0pt}{12pt}
 Discount factor($\gamma$) & 0.98\\
 \hline
 \rule{0pt}{12pt}
Training interval (T horizon) & 10 step\\
 \rule{0pt}{12pt}
 Discount factor($\gamma$) & 0.98\\
 \hline
 \rule{0pt}{12pt}
Truncating importance sampling ratio $c$ & 1.0 \\
 \hline
 \rule{0pt}{12pt}
 Optimizer & Adam \\
 \rule{0pt}{12pt}
 Learning rate& $1e-2$ \\
 \rule{0pt}{12pt}
 $e$ & $1e-8$ \\
\hline
\rule{0pt}{12pt}
Entropy term coefficient & 1e-2 \\
\rule{0pt}{12pt}
Value loss coefficient& 1 \\
\rule{0pt}{12pt}
Bias correction term & 1\\
\rule{0pt}{12pt}
Actor loss coefficient& 1 \\
\hline
\end{tabular}
\end{table}

\end{document}